\documentclass[11pt]{article}

\usepackage[top=25truemm,bottom=25truemm,left=20truemm,right=20truemm]{geometry}

\usepackage[utf8]{inputenc} % allow utf-8 input
\usepackage[T1]{fontenc}    % use 8-bit T1 fonts
\usepackage{hyperref}       % hyperlinks
\usepackage{url}            % simple URL typesetting
\usepackage{booktabs}       % professional-quality tables
\usepackage{amsfonts}       % blackboard math symbols
\usepackage{nicefrac}       % compact symbols for 1/2, etc.
\usepackage{microtype}      % microtypography

\usepackage{hyperref}
\hypersetup{
	colorlinks   = true, %Colours links instead of ugly boxes
	urlcolor     = blue, %Colour for external hyperlinks
	linkcolor    = blue, %Colour of internal links
	citecolor   = blue %Colour of citations
}

\usepackage{natbib}
\usepackage{array}
\usepackage{tabularx}
\usepackage{microtype}
\usepackage{wrapfig}
\usepackage{graphicx}
\usepackage{subfigure}
\usepackage{bm}
\usepackage{comment}
\usepackage{amsmath,amssymb,amsthm,mathtools}
\usepackage[ruled,vlined]{algorithm2e}
\usepackage{setspace}

\renewcommand{\mid}{\,:\,}

\newcommand{\black}{\textcolor{black}}
\renewcommand{\mid}{\,:\,}

\newcommand{ \chiS}{\bm{\chi}_{S}}

\newcommand{\Atin}{A_{{\bf x}_t}^{-1}}

\newcommand{\Mlin}{M(\lambda)^{-1}}

\theoremstyle{plain}
\newtheorem{theorem}     {Theorem}

\newtheorem{proposition} {Proposition}

\newtheorem{problem}     {Problem}

\newtheorem{remark} {Remark}
\newtheorem{assumption} {Assumption}

\usepackage[usenames,dvipsnames]{xcolor}
\usepackage{ifthen}

\newcommand{\argmax}{\operatornamewithlimits{argmax}}
\newcommand{\argmin}{\operatornamewithlimits{argmin}}

\newcommand{\E}{\mathbb{E}}

\newcommand{\cX}{\mathcal{X}}
\begin{document}

%\title*{\bf Selected Open Problems in Discrete Geometry and Optimization}
\title{\bf Combinatorial Pure Exploration with Full-bandit Feedback and Beyond: Solving Combinatorial Optimization under Uncertainty with Limited Observation}

% \author{Yuko Kuroki, Junya Honda, %W{\l}odzimierz Kuperberg, 
% and Masashi Sugiyama}

\author{
	Yuko Kuroki \\
	The University of Tokyo \& RIKEN AIP\\
	CENTAI Institute\\
	\texttt{yuko.kuroki@centai.eu} \\
% Yuko Kuroki\\
%  The University of Tokyo, RIKEN\\
%   \texttt{ykuroki@ms.k.u-tokyo.ac.jp} \\
  \and
Junya Honda\\
 Kyoto University \& RIKEN \\
  \texttt{honda@i.kyoto-u.ac.jp} \\
  \and
   Masashi Sugiyama\\
   RIKEN \& The University of Tokyo\\
  \texttt{sugi@k.u-tokyo.ac.jp}
}

\date{}
\maketitle

\abstract{\emph{Combinatorial optimization} is one of the fundamental research fields that has been extensively studied in theoretical computer science and operations research. When developing an algorithm for combinatorial optimization, it is commonly assumed that parameters such as edge weights are exactly known as inputs. However, this assumption may not be fulfilled since input parameters are often uncertain or initially unknown in many applications such as recommender systems, crowdsourcing, communication networks, and online advertisement. To resolve such uncertainty, the problem of \emph{combinatorial pure exploration of multi-armed bandits} (CPE) and its variants have recieved increasing attention. Earlier work on CPE has studied the semi-bandit feedback or assumed that the outcome from each individual edge is always accessible at all rounds. However, due to practical constraints such as a budget ceiling or privacy concern, such strong feedback is not always available in recent applications. In this article, we review recently proposed techniques for combinatorial pure exploration problems with limited feedback.}

% \keywords{Combinatorial optimization,  combinatorial bandits, pure exploration of multi-armed bandits.}

% \medskip

% \noindent{\bf AMS Subject Classifications (2020):} 05, 60, 62, 68, 90.



\section{Introduction}
% combinatorial actions is that 

\emph{Combinatorial optimization} is one of the fundamental research fields that has been extensively studied in theoretical computer science and operations research~\citep{Korte2012}.
Many classical problems such as the shortest path problem~\citep{dijkstra1959note}, the minimum spanning tree problem~\citep{kruskal1956shortest,prim1957shortest}, and the weighted matching problem~\citep{edmonds_1965} have been extensively discussed in such fields.
When developing an algorithm for combinatorial optimization,
it is commonly assumed that parameters such as edge weights are exactly known as inputs.
However, this assumption may not be fulfilled since input parameters are often uncertain or initially unknown in many applications such as recommender systems, crowdsourcing, communication networks, and online advertisement~\citep{Gai2012}.

\black{Over the past decade,
optimization models that are immune to data uncertainty have been studied in the field of \emph{robust optimization}~\citep{bental2009robust, Bertsimas2011}.
In the robust optimization paradigm,
uncertainty sets may often be modeled as deterministic sets such as boxes, polyhedra, or ellipsoids.
The quality of a solution is then evaluated using the realization of the uncertainty that is most unfavorable for the decision maker.
That is, robust optimization considers the worst-case scenario given uncertainty sets, and does not discuss how to design optimal sampling strategies of data to be collected.
This has led to the study of combinatorial online learning problems, i.e., \emph{combinatorial multi-armed bandits} (CMAB) problem, which combines online learning problems with combinatorial optimization and discusses how to learn unknown parameters. 
CMAB and its variants have recieved increasing attention in the online learning community~\citep{Chen2014, Chen2016matroid, Gabillon2016, Chen17a,huang2018,Cao2019, Idan2019,chen2020cpe_db,  Kuroki2020,Kuroki+19, Zhong+2020, Du+2021}.
}

\black{
The first idea of \emph{multi-armed bandits} (MAB) appeared early in the 20th century~\citep{Thompson1933}, motivated by the medical treatment design.
The popularization of MAB as a sequential decision making model would have been realized by the seminal work of \citet{Robbins1952} and~\citet{Lai1985}.
The model of MAB is described as the following learning problem.
% an agent takes an action and observes a stochastic reward as a response at each time step.
Suppose that there are $K$ possible actions, whose reward is unknown, and each action is associated with an unknown probability distribution.
An action is usually called an {\em arm}, 
which characterizes the decision making problem of the agent in the following stochastic game with discrete time steps. 
At each round $t=1, 2, \ldots$,
the agent must choose an arm to pull from $K$ arms.
When each arm is pulled, the agent can observe a stochastic reward sampled from an unknown probability distribution.
The most well-studied objective is to minimize the \emph{cumulative regret},
i.e.,
the total loss between the expected reward of the optimal arm and the expected reward of the arm collected by the agent~\citep{ bubeck2012,cesa2006prediction}.
Another popular objective is to identify the best arm,
i.e.,
the arm with the maximum expected reward among $K$ arms, by interacting with the unknown environment.
This problem,
called {\em  pure exploration} or {\em best arm identification} of the MAB,
has also received much attention recently~\citep{Even2002,Even2006,Audibert2010,Jamieson2014,Chen2015,kaufmann2016}.
This article focuses on the pure exploration problem.
}

\black{
Despite modern developments of MAB over nearly a century,
\emph{combinatorial actions} pose a challenge to these fields.
Algorithms for bandit problems typically require an action space to be small enough to enumerate, and how to deal with combinatorially large action space has been overlooked until recently in the literature.
In many real-world scenarios, indeed, our decisions are often characterized by a combinatorial structure.
For example, possible actions in real-world systems may be a subset of keywords in online advertisements~\citep{Rusmevichientong2006}, assignments of tasks to workers in crowdsourcing~\citep{Zhou2014}, or channel selection in communication networks~\citep{Huang2008}.}

\black{
The problem of combinatorial bandits is a generalization of MAB, which considers combinatorial actions; a subset of underlying arms, called a \emph{super arm}, is an action in this model, while each single arm is an action in the MAB problem.
To be more precisely, 
let us describe the setup of stochastic combinatorial bandits and feedback models with a linear objective.
Suppose that we are given a graph $G=(V,E)$ with unknown edge weights $\theta \mid E \rightarrow \mathbb{R}_+$. Each edge $e \in E$ corresponds to a \emph{base arm} and $\cX \subseteq \{0,1\}^{|E|}$ is a set of \emph{super arms} satisfying a given combinatorial structure, where each element $x \in \cX$ is an indicator vector of a super arm satisfying the given combinatorial structure such as a size-$k$ subset, path, or matching.}
At each time step $t=1,2,\ldots,T$,
an agent selects a combinatorial action $x_t \in \cX$.
Then, the agent observes random feedback depending on $x_t$ from an unknown environment.
The goal is to find the optimal action $x^* = \argmax_{ x \in \cX} \sum_{e \in E}\theta_e x_e$ with high probability.
Let $X_t(e)$ for $e \in E$ be a random variable at round $t$ independently sampled from the associated unknown distribution.
Here, the reward vector $X_t \in \mathbb{R}^{|E|}$ has its $e$-th element with $X_t(e)=\theta_e+ \eta_t(e)$, where $\theta_e$ is the expected reward and $\eta_t(e)$ is the zero-mean noise.
There are two types of feedback as follows.
\begin{itemize}
     \item [(i)]\emph{Semi-bandit feedback}: After pulling a super arm $x_t$ at round $t$, the component $X_t(e)$ for $e \in E$ is observed if and only if $x_t(e)=1$.
      \item [(ii)]\emph{Full-bandit feedback}: After pulling a super arm $x_t$ at round $t$, only the sum of rewards $x_t^\top X_t$ is observed.
\end{itemize}

Most prior work for combinatorial pure exploration assumed that the outcome from each base arm is always accessible at all rounds~(e.g.~\citep{Chen2014, Chen+16, Chen2016matroid, Gabillon2016, Jun+16,Chen17a,huang2018,Cao2019, Jourdan21}).
However, in most application domains, such strong feedback is not always available,
since it is costly to observe
a reward of individual arms, 
or sometimes we cannot access feedback from individual arms.
For example, in crowdsourcing,
we often obtain a lot of labels given by crowdworkers,
but it is costly to compile labels according to labelers.
In the case of transportation networks, it is not easy to observe a delay in each section of a path due to some system constraints.
In social networks, due to some privacy concerns and data usage agreements, 
it may be impossible even for data owners to obtain the estimated number of messages exchanged by two specific users.
\black{
To overcome these issues, 
\citet{Kuroki+19} studied \emph{combinatorial pure exploration with full-bandit linear feedback} (CPE-BL) problem, and proposed a non-adaptive
algorithm to solve this problem. 
Later, \cite{Idan2019} proposed an adaptive combinatorial successive acceptance and rejection algorithm. These algorithms work for the top-$k$ case.  \citet{Kuroki2020} also studied the specific graph optimization problem, called the densest subgraph problem and its pure exploration problem with full-bandit feedback. \citet{Du+2021} further extended these studies by proposing the first adaptive polynomial-time algorithm for full-bandit and static algorithm for partial-linear feedback under general combinatorial constraints.
}

\black{
We note that in the literature of stochastic combinatorial bandit for regret minimization, most existing work has considered semi-bandit feedback (e.g.~\citep{chenwei13, kveton15,ChenJMLR17,Zheng2017,pierre2019}).
There are only a few studies dealing with the full-bandit feedback even for the top-$k$ case~\citep{Idan2019, Agarwal2021}. Adversarial cases, in which an adversary controls the arms and tries to defeat the learning process, have also been studied in the literature~\citep{Abernethy+2008, Cesa2012, Combes2015, Rad2021}. Some of them can deal with bandit feedback and nonlinear reward, but regret minimization algorithms for adversarial cases cannot be applied to solve the pure exploration problem. 
}

\black{Why is dealing with such limited feedback in combinatorial pure exploration so hard?
Stochastic bandit problems have been analyzed by two different approaches:
a frequentist approach, where the parameter is a deterministic unknown quantity, and a Bayesian approach, where the parameter is drawn from a prior distribution.
The vast majority of combinatorial and linear bandit work follows a frequentist approach.
They often employ the optimism principle exemplified by \emph{upper confidence bound} (UCB) algorithm~\citep{yadkori11,Chen+16}, which uses the data observed so far to assign to each arm a value, called the UCB that is an overestimate of the unknown mean with high probability.
To handle full-bandit feedback, the frequentist approach may rely on the least-square estimator, and the combinatorial structure results in a confidence region in the form of an ellipsoid.
With a confidence ellipsoid, algorithms often require complex optimization for determining the next arm to pull or stopping conditions, which might involve quadratic optimization with combinatorial constraints (as to be discussed in Section~\ref{sec:static_algorithm}).}

\black{
One might think that algorithms for linear bandits can deal with combinatorial action spaces.
Most linear bandit algorithms have, however, the time complexity at least proportional to the number of arms~\citep{Dani2008, yadkori11,Kwang2017} since all the existing linear bandit algorithms execute a brute-force search to solve such a kind of optimization problem.
Therefore, a naive use of them is computationally infeasible since the number of actions $K$ is exponential.
Therefore, we need to develop different techniques to deal with full-bandit feedback in combiantorial settings.}

\black{In this article, we review the formulation, technique, and sample complexity results for combinatorial pure exploration with limited feedback by introducing the work of \citet{Kuroki+19} and \citet{Du+2021}.}

\section{Formulations of Combinatorial Pure Exploration with Full-bandit Feedback and Related Problem}

In this section,
we provide the formulation of the \emph{combinatorial pure exploration with (full-)bandit linear feedback} problems (CPE-BL).
We also introduce the problem of \emph{best arm identification in linear bandits} (BAI-LB) as a related problem.

\subsection{Combinatorial Pure Exploration with Full-bandit Feedback}
In the \emph{multi-armed bandits} (MAB), 
an action corresponds to a single arm.
 On the other hand,
 in the \emph{combinatorial bandits},
 given a set of \emph{base arms} $[d]=\{1,2,\ldots,d\}$ for integer $d$,
 each action is a set of base arms, called the \emph{super arm}.
In the problem of CPE-BL,
an agent samples a super arm $x_t \in \cX$ at any round $t$, where $\cX \subseteq \{0,1\}^d$ is a family of super arms rather than a set of base arms $[d]$. 
%In our model, we address two types of feedback: full-bandit  and partial-linear feedback.
In the full-bandit setting, the agent can only observe the sum of rewards $r_{x_t}= x_t^{\top}(\theta+\eta_t)$ at each pull, where each element $\eta_t(e)$ of noise vector $\eta_t$ is a zero-mean random variable.

Let ${\tt Out} \in \cX$  be an output of an algorithm and $x^*$ be the optimal arm with the highest expected reward, i.e., $x^*=\argmax_{x \in \cX}x^{\top} \theta$.
The goal is to find ${\tt Out} \in \cX$  while guaranteeing ${\tt Out}=x^*$ with high probability. 
In the literature~\citep{Audibert2010,Bubeck+2009,Gabillon2011},
there are two different settings: the {\em fixed confidence} and {\em fixed budget} settings defined as follows.
\begin{itemize}

 \item \emph{Fixed confidence setting}:
 The agent can determine when to stop the game.
After the game is over, she needs to report ${\tt Out} \in \cX$ satisfying $\Pr[{\tt Out}=x^*] \geq 1-\delta$ for given confidence parameter $\delta \in (0,1)$.
 The agent's performance is evaluated by her {\em sample complexity}, i.e., the number of pulls used by her in the game.

 \item \emph{Fixed budget setting}:
The agent needs to minimize the {\em probability of error} $\Pr[{\tt Out} \neq x^*]$ within a fixed number of rounds $T$.
\end{itemize}
In this article, we mainly focus on the fixed confidence setting.
% In Section~\ref{sec:dense}, we also address the fixed budget setting %Chapter~\ref{chap:dense},
% where the quality of output is not necessarily optimal but a $1/2$-approximate solution.
In the fixed confidence setting,
an algorithm that satisfies $\Pr[{\tt Out}=x^*] \geq 1-\delta$ for given $\delta \in (0,1)$ is called \emph{$\delta$-probably approximately correct} (PAC).
Any algorithms for the fixed confidence setting consist of the following three components:

\begin{itemize}
    \item [1. ]A \emph{stopping rule:} which controls when the agent stops the sampling procedure for data acquisition.
    %, and depend on past arms selection and observations;
    \item [2. ]A \emph{sampling rule:} which determines, based on past observations, which arm $x_t$ is chosen at round $t$. 
    \item [3. ]A \emph{recommendation rule:} which chooses the arm from $\cX$ that is to be reported as the optimal arm.
\end{itemize}
The summary of a general procedure of CPE-BL is given in Algorithm~\ref{alg:pre}.
\begin{algorithm}[t!]
%\caption{Ellipsoid Confidence Bound with First-Order Approximation ({\tt ECB-FOA)}}
\caption{A general procedure of combinatorial pure exploration problems with full-bandit feedback}

	%\caption{ECB with Additive Approximation	}
	\label{alg:pre}
	\SetKwInOut{Input}{Input}
	\SetKwInOut{Output}{Output}
	\Input{ Confidence level $\delta \in (0,1)$, a set of base arms $[d]$}

	\While{a stopping rule is $\mathrm{False}$}{
    $t \leftarrow t+1$;
    
    Pull a super arm $x_t \in \cX$ by a sampling rule;
    
    %uniformly or based on~\eqref{G_allocation};
    Observe the sum of random rewards $r_{x_t}= x_t^{\top}(\theta+\eta_t)$;
    
    Update statistics;
}
    
    \Return{a super arm $\mathrm{Out}$ by a recommendation rule.}
\end{algorithm}

\subsection{Related Problem: Best Arm Identification in Linear Bandits}
In this section,
we introduce the problem of \emph{best arm identification in linear bandits} (BAI-LB) as a related problem.

\citet{Auer03} first introduced the {\em linear bandit}, an important variant of MAB.
In the linear bandit,
for a dimension $d>0$, an agent has the set of arms $\cX \subseteq \mathbb{R}^d$.
The expected reward for arm $x \in \cX$ is written by $x^{\top}\theta$,
where $\theta \in \mathbb{R}^d$ is an unknown parameter.
It should be noted that the linear bandit is a generalized model of the ordinary MAB and combinatorial bandits with linear objectives.
When $\cX=\{e_1,e_2,\ldots,e_d\}$ 
where $e_1,e_2,\ldots,e_d$
are the standard basis of the $d$-dimensional Euclidean space,
%are the unit vectors in the standard Euclidean basis,
the linear bandit is reduced to the ordinary MAB.
When $\cX \subseteq \{0,1\}^d$ represents a combinatorial action set, 
the linear bandit coincides with 
the combinatorial bandits with a linear reward function.

In BAI-LB, at each round $t$,
the agent pulls an arm $x_t \in \cX$, and then observes a reward
$r_t = x^{\top}\theta + \eta_t$, where $\eta_t$ is a zero-mean random variable.
The goal is to identify the best arm with the highest expected rewards.
\citet{Soare2014} addressed BAI-LB in the fixed confidence setting and first provided a static allocation algorithm for BAI-LB, whose sampling rule is independent on any past observation.
For its design, \citet{Soare2014} introduced the connection between BAI-LB and the \emph{G-optimal experimental design}~\citep{pukelsheim2006}.
Since then, there has been a surge of interest
in BAI-LB~\citep{Degenne+2020,Fiez2019,jedra2020optimal, Karin16,Katz+2020,Tao2018, Xu2018,zaki2019towards,Zaki+2020}. 
%\begin{landscape}
%\centering

\begin{table*}[t]
	\centering
	\caption{Sample complexity results for CPE-BL and BAI-LB. ``General'' represents that the algorithm works for combinatorial structures including size-$k$ subsets, paths, matchings, and matroids. $\tilde{O}(\cdot)$ only omits $\log \log$ factors. Some specific notation are given in the footnote. This table is a slight modification of Table~1 in \citet{Du+2021}.
	}\label{table:resultsfull}
	\renewcommand\arraystretch{1.8}
	\scalebox{0.68}{
	\begin{tabular}{|c|c|c|c|c|c|}
		\hline
		Reference &Sample complexity\footnotemark[1]&Case&Problem Type&Strategy&Time\\
		\hline
% 		{\bf 
% 		\textsf{GCB-PE} (ours, Thm.~\ref{thm:GCB_ub})} &$O \big( \frac{ |\sigma| \beta_{\sigma}^2 L_p^2}{\Delta_{\textup{min}}^2} \log \frac{ \beta_{\sigma}^2 L_p^2}{\Delta_{\textup{min}}^2\delta} \big)$& General &CPE-PL&Static&$\mathrm{Poly}(d)$\\
% 		\hline
		\citep{Du+2021} &$\tilde{O} \big(\sum_{i=2}^{\lfloor \frac{d}{2} \rfloor} \frac{1}{\Delta_i^2} \log \frac{|\mathcal{X}|}{\delta}    +\frac{d^2 m \xi_{\max}({\widetilde{M}({\lambda})}^{-1})}{ \Delta^2_{d+1}} \log \frac{|\mathcal{X}|}{\delta} \big)$&General &CPE-BL& Adaptive  &$\mathrm{Poly}(d)$\\
% 		\hline
% 		\textsf{ICB} in Chapter~\ref{chap:topk}&$\tilde{O}\big(\frac{d  \xi_{\max}({M({\lambda})}^{-1})\rho(\lambda)}{\Delta^2_{\min}} \log \frac{d  \xi_{\max}({M({\lambda})}^{-1}) \rho(\lambda)}{\Delta^2_{\min}\delta}\big)$ &General\footnotemark[2] &CPE-BL& Static&$\mathrm{Poly}(d)$\\
		\hline
		 \citep{Kuroki+19} &$\tilde{O}\big(\frac{d^{1/4}k\xi_{\max}({M({\lambda})}^{-1})\rho(\lambda)}{\Delta_{\min}^2} \log \frac{d^{1/4}k\xi_{\max}({M({\lambda})}^{-1})\rho(\lambda)}{\Delta_{\min}^2  \delta}\big)$ &Top-$k$ &CPE-BL& Static&$\mathrm{Poly}(d)$ \\
		\hline
		\citep{Idan2019} &$\tilde{O}\big(\sum^{d}_{i=2}\frac{1}{\tilde{\Delta}_i^2} \log\frac{d}{\delta}\big)$&Top-$k$ &CPE-BL&  Adaptive &$\mathrm{Poly}(d)$ \\
		\hline
	\citep{Soare2014}&$O\big(\frac{d}{\Delta_{\textup{min}}^2}\log \frac{|\cX|}{\delta \Delta^2_{\min}} +d^2\big)$&$\cX \subseteq \mathbb{R}^d$&BAI-LB& Static& $\Omega(|\cX|)$ \\
		\hline
\citep{Karin16}&$O\big(\frac{d}{\Delta_{\textup{min}}^2} \log \frac{|\cX|}{\delta \Delta_{\min}}+ d\log \delta^{-1}\big)$&$\cX \subseteq \mathbb{R}^d$&BAI-LB& Static& $\Omega(|\cX|)$\\
		\hline
\citep{Xu2018}&  $\tilde{O}\big(d  \sum_{x \in \cX} H_x\log \frac{d|\cX|}{\delta} \cdot \sum_{x \in \cX} H_x\big)$ &$\cX \subseteq \mathbb{R}^d$&BAI-LB& Adaptive & $\Omega(|\cX|)$\\
% 		\hline
% 		$\mathcal{Y}$-\textsf{ElimTil}~\citep{Tao2018}&$\tilde{O}\big(\frac{d}{\Delta_{\textup{min}}^2} (\log \delta^{-1} + \log |\mathcal{X}|)\big)$&$\cX \subseteq \mathbb{R}^d$&BAI-LB& Adaptive& $\Omega(|\cX|)$\\
		\hline
\citep{Tao2018}&$\tilde{O}\big(\sum^{d}_{i=2}\frac{1}{\Delta_i^2}(\log \delta^{-1} + \log |\mathcal{X}|)\big)$&$\cX \subseteq \mathbb{R}^d$&BAI-LB& Adaptive & $\Omega(|\cX|)$\\
		\hline
	\citep{Fiez2019}&$O \big(\sum^{\left \lfloor \log_2(4/\Delta_{\min})  \right \rfloor}_{t=1} 2(2^t)^2\tilde{\rho}(\mathcal{Y}(S_t)) \log(t^2|\mathcal{X}|^2/\delta) \big)$&$\cX \subseteq \mathbb{R}^d$&BAI-LB& Adaptive& $\Omega(|\cX|)$\\
		\hline
\citep{Degenne+2020}&$\mathop{\lim \sup}_{\delta \rightarrow 0} \frac{\mathbb{E}_{\theta}[\tau_{\delta}]}{\log(1/ \delta)} \leq \min_{\lambda \in \triangle(\cX)}\max_{x\in \mathcal{\cX} \setminus\{x^*\}} \frac{2||x^*-x||_{M(\lambda)^{-1}}^2}{((x^*-x)^{\top}\theta)^2} $&$\cX \subseteq \mathbb{R}^d$&BAI-LB& Adaptive& $\Omega(|\cX|)$ \\
		\hline
\citep{Katz+2020}&$O\big( \big(\min_{\lambda \in \triangle(\cX)}\max_{x\in \mathcal{\cX} \setminus\{x^*\}} \frac{||x^*-x||_{M(\lambda)^{-1}}^2}{((x^*-x)^{\top}\theta)^2} + \gamma^*\big)  \log(1/ \delta)\big)$&$\cX \subseteq \mathbb{R}^d$&BAI-LB& Adaptive& $\Omega(|\cX|)$ \\
		\hline
% 		Lower Bound \citep{Fiez2019}&$\mathbb{E}_{\theta}[\tau_{\delta}] \geq \min_{\lambda \in \triangle(\cX)}\max_{x\in \mathcal{\cX} \setminus\{x^*\}} \frac{||x^*-x||_{M(\lambda)^{-1}}^2}{((x^*-x)^{\top}\theta)^2} \log(1/2.4 \delta)$&$\cX \subseteq \mathbb{R}^d$&BAI-LB& - & - \\
% 		\hline
	\end{tabular}
	}
\end{table*}
\footnotetext[1]{
	%	The sample complexity of \textsf{LinGame(-C)} is an asymptotic result.
	Notation appearing in the table but not relevant in our problem setting are given below:
	$\rho(\lambda)=\max_{x \in \mathcal{X}}\|x\|^2_{M(\lambda)^{-1}}$.
	$\tilde{\Delta}_i=\theta_i - \theta_{k+1}$ if $i \leq k$ and $\theta_k-\theta_i$ otherwise.
	$H_x=\underset{x_i,x_j \in \cX}{\max}\frac{\bar{\rho}_x(x_i,x_j)}{\max\{\bar{\Delta}^2_i \bar{\Delta}^2_j\}}$  where $\bar{\Delta}=(x^*-x_i)^{\top}\theta$ if $x_i \neq x^*$, $\argmin_{x \in \cX} x^*-x$ otherwise, and $\bar{\rho}_x(x_i,x_j)$ is a term defined by the optimal solution to a convex optimization (see (11) in~\citet{Xu2018}).
	$S_t=\{x \in \cX \mid (x^*-x)^{\top}\theta\leq 4 \cdot 2^{-t} \}$. $\mathcal{Y}(S_t)=\{ x-x' \mid \forall x,x' \in S_t, x \neq x'\}$.
	$\tilde{\rho}(\mathcal{Y}(S_t))=\min_{\lambda \in \triangle(\cX)}\max_{v \in \mathcal{Y}(S_t) }\|v\|_{\Mlin}$. 
	$\gamma^*=\min_{\lambda \in \triangle(\cX)} \mathbb{E}_{\eta \sim N(0,1)} \left[ \max_{x\in \mathcal{\cX} \setminus\{x^*\}} \frac{(x^*-x)^\top M(\lambda)^{-1/2} \eta}{(x^*-x)^{\top}\theta} \right]^2. $
}

%  \footnotetext[2]{\textsf{ICB} runs in polynomial time for combinatorial constraints such that $\mathrm{P}_1$ in~\eqref{ICB_opt} is solved in polynomial time.}

 While the existing BAI-LB algorithms achieve satisfactory sample complexity, none of them can solve CPE-BL in polynomial time in $d$ since they implicitly assume that $|\cX|$ is small enough to enumerate.
However, some techniques established in the literature of linear bandits can be used in order to deal with full-bandit feedback in combinatorial settings. For example, \citet{Kuroki2020,Kuroki+19} uses the least-squares estimator for unknown vector $\theta$ and a high probability bound proposed in \citet{yadkori11,Soare2014}.
\citet{Du+2021} uses the randomized estimator proposed in \citet{Tao2018} and invoke their algorithm as a subroutine.
\black{Sample complexity bounds of some of these studies are summarized in Table~\ref{table:resultsfull}.}

\subsection{Notation}
We introduce some notation used in this article.
For $S \subseteq [d]$, we use $\chiS$ to denote an indicator vector with the $i$-th coordinate 1 for $i \in S$ and 0 otherwise.
%For a matrix $A \in \mathbb{R}^{d \times d}$, 
%$A(i,j)$ denotes the $(i,j)$-th entry of $A$ for $i,j \in [d]$.
For a vector $x \in \mathbb{R}^d $ and a matrix $A\in \mathbb{R}^{d\times d}$, let $\|x\|_A=\sqrt{x^\top Ax}$. 
%For a vector $\theta \in \mathbb{R}^d$
%and a set $S \subseteq [d]$, 
%we define $\theta(S)  =\sum_{e\in S} \theta(e)$.
For a given set $\mathcal{X}$, we use $\triangle(\mathcal{X})$ to denote the family of probability distributions over $\mathcal{X}$.
%\paragraph{Matrices.}
%We call a symmetric matrix $A \in \mathbb{R}^{d \times d}$ positive semidefinite if $x^{\top}Ax \geq 0$ for all $x \in \mathbb{R}^d$ and we call $A$ positive definite if $x^{\top}Ax > 0$ for all $x \neq \bm{0}$.
%For any $x \in \mathbb{R}^d$, let ${\rm Diag}(x)$ be the diagonal matrix whose $i$-th diagonal component is $x(i)$ for $i \in [d]$.
We let $A^+$ denote the Moore-Penrose pseudoinverse of $A$.
We let $\xi_{\max}(A)$ and $\xi_{\min}(A)$ be the maximum and minimum eigenvalues of $A$, respectively.
%The condition number of matrix $A$ is defined by $\xi_{\max}(A)/\xi_{\min}(A)$.
The identity matrix is denoted by $I$ or,
when we want to stress its dimension $d$, by $I_d$.
For distribution $\lambda \in \triangle(\mathcal{X})$ for a finite set $\cX$, we define  $\mathrm{supp}(\lambda)=\{ x \mid \lambda(x) > 0\}$, $M(\lambda)=\mathrm{E}_{z \sim \lambda}[z z^{\top}]$
and ${\widetilde{M}({\lambda}})=\sum_{x \in \mathrm{supp}(\lambda)}x x^{\top}$.

%\paragraph{Graphs.}
A graph $G=(V,E)$ consists of a finite nonempty set $V$ of vertices and finite set $E$ of edges. For a subset of vertices $S\subseteq V$, let $G[S]$ denote the subgraph induced by $S$, i.e., $G[S]=(S,E(S))$ where $E(S)=\{\{u,v\}\in E\mid u,v\in S\}$.
% Two vertices that are joined by an edge are called adjacent or neighbors.
% For $S \subseteq V$ and $ v \in S$, let $N_S(v) =\{u \in S \mid \{u,v\} \in E \}$ be the set of neighboring vertices of $v$ in $G[S]$ and let $E_S(v) = \{\{u, v\}\in E \mid u\in N_S(v) \}$ be the set of incident edges to $v$ in $G[S]$.

\section{Static Algorithm}\label{sec:static_algorithm}

In order to handle full-bandit (linear) feedback,
the \emph{least-squares estimator} is used to estimate the unknown vector $\theta \in \mathbb{R}^d$.
Let $\textbf{x}_t= (x_1,\ldots,x_t)$ be the sequence of super arm selections,
and $(r_{x_1}, \ldots, r_{x_t}) \in \mathbb{R}^{t}$ be the corresponding sequence of observed rewards for time step $t$.
Given $\textbf{x}_t$, an unbiased least-squares estimator for $\theta \in \mathbb{R}^d$
can be obtained by
\begin{align}\label{ols}
\widehat{\theta}_t
=A_{{\bf x}_t }^{-1}b_{{\bf x}_t}\in \mathbb{R}^{d}, 
\end{align}
where 
\begin{align}\label{statics}
A_{{\bf x}_t}= \sum_{t'=1}^t x_{t'}x_{t'}^{\top}  \in \mathbb{R}^{d \times d}
\ \ \text{and} \ \
b_{{\bf x}_t}=\sum_{t'=1}^t x_{t'} r_{x_{t'}} \in \mathbb{R}^d.
\end{align}
It suffices to consider the case where  $A_{{\bf x}_t}$ is invertible,
since we shall exclude a redundant feature when any sampling strategy cannot make $A_{{\bf x}_t}$ invertible.

\subsection{Ellipsoidal Confidence Bound and Computational Hardness}\label{sec:confidence_bound}
%\subsection{Computational Hardness}\label{sec:hardness}
% We shall assume that
For any $\cX \subseteq \mathbb{R}^d$ and ${\bf x}_t$ \emph{fixed beforehand},
% Soare, Lazaric, and Munos~\cite{Soare2014}
%Soare et al.~\cite{Soare2014}
\citet{Soare2014}
provided the following proposition on the confidence ellipsoid for ordinary least-square estimator $\widehat{\theta}_t$.
% In CPE-BL,
% if we assume that each base arm follows R/
% the proposition holds for $\sigma=kR$.

%\begin{proposition}[\citep[Proposition~\ref{proposi_static}]{Soare2014}]
\begin{proposition}[{\citet[Proposition~1]{Soare2014}}]\label{proposi_static}
Let $\epsilon_t$ be a noise variable
bounded as $\epsilon_t \in [-\sigma, \sigma]$ for $\sigma >0$.
Let $c=2\sqrt{2}\sigma$ and $c'=6/\pi^2$ and fix $\delta \in (0,1)$.
Then, for any fixed sequence ${\bf x}_t$,
with probability at least $1-\delta$,
the inequality
\begin{align}\label{ineq:ellipsoid}
|x^{\top}\theta-x^{\top} \widehat{\theta}_t| \leq C_t \|x\|_{\Atin}
\end{align}
holds for all $t \in \{ 1,2,\ldots\}$
and $x \in \mathbb{R}^d$,
where $C_t =c\sqrt{\log(c't^2 |\cX|/\delta) }$.
\end{proposition}

In any algorithms for the fixed confidence setting,
the agent continues sampling a super arm
until a certain stopping condition is satisfied.
In order to check the stopping condition,
existing algorithms for BAI-LB involve the following confidence ellipsoid maximization (CEM) to obtain the most uncertain super arm:
\begin{align}\label{prob:CEM}
\text{CEM:}\quad \text{maximize} \  \|x \|_{\Atin} \ \ \text{subject to}\   x \in \cX,
%&            &    &              &  &x_i \in \{0,1\} \ \ &\forall i \in [d], \notag
\end{align}
%\begin{alignat}{4}\label{prob:CEM}
%&\text{CEM:}&\ \ &\text{max.}   &\ &  \|\chiM \|_{\Atin} \\\notag
%&            &    &\text{s.t.}   &  &M \in \cM,
%%&            &    &              &  &x_i \in \{0,1\} \ \ &\forall i \in [d], \notag
%\end{alignat}
where recall that $\|x \|_{\Atin}= \sqrt{x^{\top} \Atin x}$.
Most existing algorithms in linear bandits
implicitly assume that an optimal solution to CEM can be exhaustively searched.
%employ an {\em exhaustive} search
%\citep[e.g., ][]{Soare2014, Xu2018}.
However,
since the number of super arms $|\cX|$ is exponential with respect to the input size in combinatorial settings,
it is computationally intractable to exactly solve it.
Let $W \in \mathbb{R}^{d \times d}$ be a symmetric matrix.
When we consider size-$k$ subsets as combinatorial structures in CPE-BL,
CEM introduced in~\eqref{prob:CEM} can be naturally represented by the following 0-1 quadratic programming problem:
\begin{alignat}{4}\label{prob:QM}
&\text{QP:}&\ \ &\text{maximize}   &\ &  \sum_{i=1}^d\sum_{j=1}^d w_{ij}x_i x_j \\\notag
&            &    &\text{subject to}   &  &\sum_{i=1}^d x_i=k,\\\notag
&&&&&x_i \in \{0,1\}, \ \ &\forall i \in [d].\\\notag
%&            &    &              &  &x_i \in \{0,1\} \ \ &\forall i \in [d], \notag
\end{alignat}
%Since QP is known to be NP-hard~\cite{Taylor2016},
Notice that QP can be seen as an instance of the {\em uniform quadratic knapsack problem}, which is known to be NP-hard~\citep{Taylor2016}, and there are few results of polynomial-time approximation algorithms even for a special case.

Therefore,
we need approximation algorithms for CEM or
a totally different approach 
for solving CPE-BL, since the solution may involve a computational hard optimization if we naively use similar algorithms in linear bandits.

\subsection{Polynomial-time Static Algorithm and Sample Complexity}\label{sec:topk_bandit_algorithm}

To cope with computational issues, \citet{Kuroki+19} first designed a polynomial-time approximation algorithm\footnote{An $\alpha$-approximation algorithm for a maximization problem is a polynomial-time algorithm that finds a feasible solution whose objective value (OBJ) is within a ratio $\alpha$ of the optimal value (OPT), i.e., $\mathrm{OBJ} \geq \alpha \times \mathrm{OPT}$.} for a 0-1 quadratic programming problem to obtain the maximum width of a confidence ellipsoid.
By utilizing algorithms for
a classical combinatorial optimization problem,
called the
% In the design of the algorithm, 
% we utilize algorithms
% for a classical combinatorial optimization problem
% called the 
{\em densest $k$-subgraph problem (D$k$S)}~\citep{Asahiro2000,Bhaskara+10,Feige2001},
they designed an approximation algorithm that admits theoretical performance guarantee for QP in \eqref{prob:QM} with positive definite matrix $W$.
The current best approximation result for the D$k$S has an approximation ratio of $\mathrm{\Omega}(1/|V|^{{1/4}+\epsilon})$ for any $\epsilon >0$~\citep{Bhaskara+10}.
Therefore, we have the following theorem.
% The definition of the D$k$S is as follows.
% Let $G=(V,E,w)$ be an undirected graph with nonnegative edge weight $w=(w_e)_{e\in E}$. 

\begin{theorem}[{\citet[Theorem~1]{Kuroki+19}}]\label{thm:approx}
For QP with any positive definite matrix $W\in \mathbb{R}^{d \times d}$,
there exists an $\mathrm{\Omega}\left(\frac{1}{(k-1)d^{{1/4}}}
\frac{\xi_{\min}(W)}{\xi_{\max}(W)}
\right)$-approximation algorithm.
%where $r$ is $\min_{C \in {\bf C}_t }  T_{C}(t) / t $.
\end{theorem}

Based on their approximation algorithm,
they proposed the first bandit algorithms for the top-$k$ case, that runs in $O(\log |\cX|)$ time
and provided an upper bound of the sample complexity
which is still worst-case optimal.
Their algorithm employs a static continuous allocation $\lambda \in \triangle(\cX)$ which is independent of any past observation and fixed before the agent starts the stochastic game.
To check the stopping condition, they approximately solved CEM to obtain the most uncertain super arm to guarantee that the current empirically best super arm $\hat{x}^*=\argmax_{x \in \cX}x^{\top}\hat{\theta}_t$ is indeed the optimal super arm.

Let us define the {\em minimum gap} as $\Delta_{\min}=\argmin_{x \in \cX \setminus \{ x^*\}} (x^*-x)^{\top}\theta$.
We define $\Lambda_{\lambda}= \sum_{x \in \cX}\lambda(x)  x x^{\top}$ as a (continuous) design matrix, and define the problem complexity $H$ as
\[H =\frac{\rho(\lambda)}{\Delta_{\min}^2},\]
where $\rho(\lambda)= \max_{x \in \cX} \| x \|_{\Lambda^{-1}_{\lambda}}^2$,
which also appeared in~\citet{Soare2014}.
Then, we have the following theorem.
% The next theorem shows that
% {\tt SAQM} is $(\varepsilon ,\delta)$-PAC
% %for the top-$k$ arm identification with full bandit feedback,
% and gives a problem-dependent sample complexity bound.

\begin{theorem}[{\citet[Theorem~3]{Kuroki+19}}] \label{thm:topk}
Given any instance of CPE-BL,
with probability at least $1-\delta$, 
if we have an $\alpha$-approximation algorithm for CEM,
there exists an algorithm that
returns
an optimal super arm $x^*$,
and the total number of samples $T$ is bounded as follows:
\[
T \leq  8\left(3+\frac{1}{\alpha}\right)^2 \sigma^2 H \log\left (   \frac{|\cX|}{\delta} \right)+C(H, \delta),
\]
where %$C(H, \delta)$ is denoted as 
\begin{align*}
C(H, \delta)&=O\left(\sigma^2 H \log  \left(   \frac{\sigma^2}{\alpha^2}H +\log\left( \frac{|\cX|}{\delta}\right)   \right)\right)\notag.
\end{align*}
\end{theorem}
For the top-$k$ arm identification setting,
we have $\alpha= \mathrm{\Omega}\left( k^{-\frac{1}{2}}d^{-\frac{1}{8}} \sqrt{\frac{\xi_{\min}(\Lambda_{\lambda})}{\xi_{\max}(\Lambda_{\lambda})}} \right)$  in Theorem~\ref{thm:topk}.
%For the MAB, we have $\alpha=1$ because CEM can be exactly solved in polynomial time since $A_t$ becomes a diagonal matrix at any round $t>n$.

\section{Adaptive Algorithm}
In the previous section,
we have discussed a static algorithm, which has a heavy dependence on minimum gap $\Delta_{\textup{min}}$ in the sample complexity and empirically requires a large number of samples for instances with small $\Delta_{\textup{min}}$ (see \citet{Kuroki+19} for detailed experimental results).
\citet{Idan2019} developed a polynomial-time adaptive algorithm \textsf{CSAR} but it only works for the top-$k$ case. 
To resolve such a drawback,
\citet{Du+2021} designed the first {\em polynomial-time adaptive} algorithm for general combinatorial structures, whose sample complexity matches the lower bound (within a logarithmic factor) for a family of instances and has a mild dependence on the minimum gap $\Delta_{\textup{min}}$.

\subsection{Randomized Least-Squares Estimator}

Let us consider the randomized least-square estimator defined by~\citet{Tao2018}.
Let $y_1, \ldots, y_n \in \cX$ be $n$ $i.i.d.$ samples following a given distribution $\lambda \in \triangle(\cX)$, and let the corresponding rewards be $r_1, \ldots,r_n$ respectively. Let $b=\sum_{i=1}^n y_i r_i \in \mathbb{R}^d$.
Then, the randomized estimator $\hat{\theta}$ is given by
\[\hat{\theta}=A^{-1}b \in \mathbb{R}^d,
\]
where $A=n M(\lambda) \in \mathbb{R}^{d \times d}$ (recall that $M(\lambda)=\mathrm{E}_{z \sim \lambda}[z z^{\top}]$).
Based on this randomized least-squares estimator, \citet{Tao2018} proposed an algorithm, named \textsf{ALBA}, for BAI-LB, which is an elimination-based algorithm, where in round $q$ it 
identifies the top $d/2^q$ arms and discards the remaining arms.
Their algorithm runs in time polynomial to $|\cX|$, and thus it cannot be applied to CPE-BL since $|\cX|$ is exponential to the instance size in combinatorial problems.

\subsection{Two-phased Algorithm and Improved Sample Complexity}
As a remedy for the computational issues, \citet{Du+2021} proposed a polynomial-time algorithm, namely \textsf{PolyALBA}, in which we have two phases: the first phase is for finding a polynomial-size set of super arms which contains the optimal super arm with high probability and discards other super arms. The second phase focuses on sampling super arms among the rest by adaptive elimination procedures.

\begin{figure}[t]
    \centering
    \includegraphics[width=0.45\textwidth]{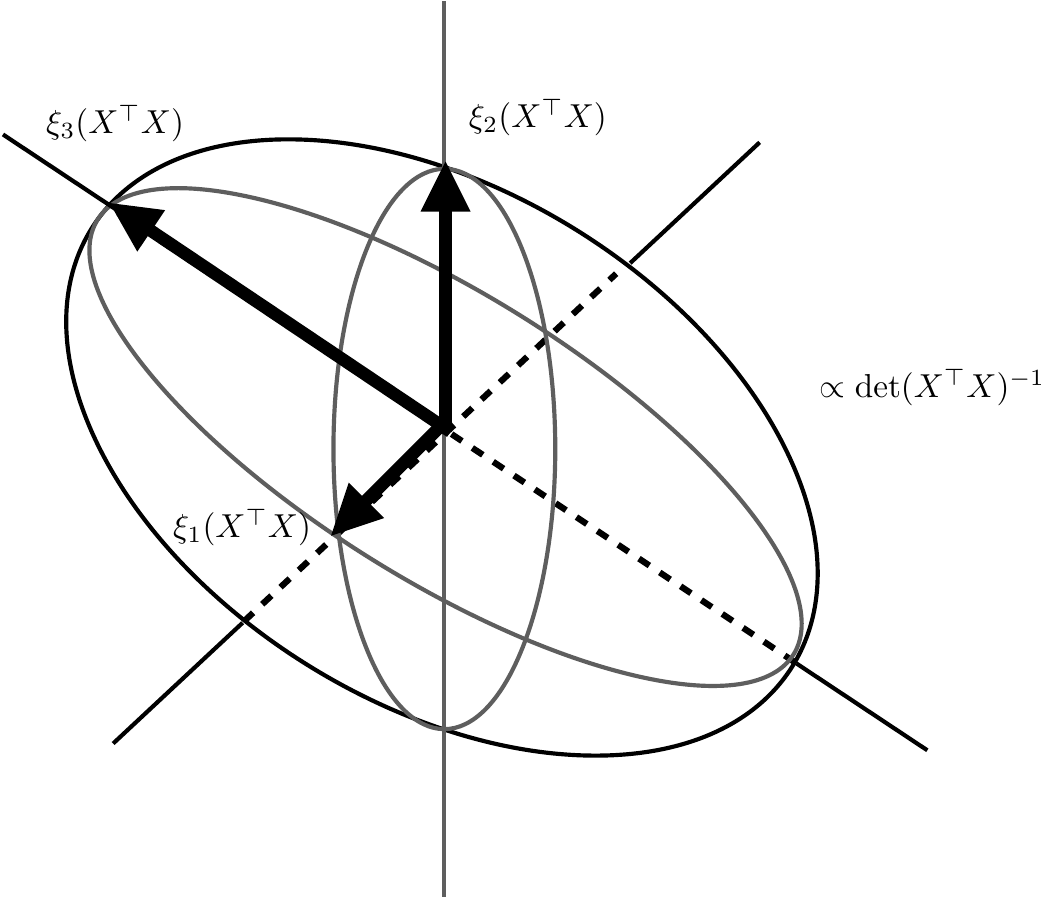}
    \caption{The illustration of a confidence ellipsoid, where $X=(x_1, \ldots, x_d)$ is the model matrix. The shape of the ellipsoid depends on the information matrix $X^{\top}X$.}
    \label{fig:ellipsoid}
\end{figure}

Some of the BAI-LB algorithms (e.g.,~\citet{Soare2014,Tao2018} and \citet{Fiez2019}) require solving the following G-optimal design problem for their design:
\[
 \min_{\lambda \in \triangle(\cX)}  \max_{x \in \mathcal{X}} \|x\|_{M(\lambda)^{-1}},
\]
where $M(\lambda)^{-1}=(\sum_{x \in \mathcal{X}} \lambda(x) x x^{\top})^{-1}$ is the error covariance matrix.
The G-optimal design aims to minimize the maximum prediction variance, and as shown in \citet{kiefer_wolfowitz_1960}, the continuous G-optimal design and D-optimal design are equivalent when the errors are homoscedastic.
Geometrically, this corresponds to designing the experiment to minimize the volume of the resulting confidence ellipsoid~\citep{boyd2004} (see Figure~\ref{fig:ellipsoid}).
% A design which minimizes the maximum of the standardized variance function
% % over the design region X is called a G-optimal design.
% As shown in \citet{kiefer_wolfowitz_1960}, 
% G-optimal design
%An-optimal design aims to minimizes the average prediction variance over design space
The above G-optimal design problem is hard to compute when $|\mathcal{X}|$ is exponentially large, since we have an exponential number of variables and the inner optimization is already hard to solve as discussed in the previous section.
%Then, the proposed algorithm  avoids this issue by selecting a set of super arms.
To avoid the high computation cost, the algorithm \textsf{PolyALBA} proposed in \citet{Du+2021}
chooses \black{any polynomially sized} set of super arms $\cX' \subseteq \mathcal{X}$.
Then we can obtain 
\[
\lambda^*_{\cX'}=\argmin_{\lambda \in \triangle(\cX' )} \max_{x \in \cX'} \|x\|_{M(\lambda)^{-1}}.
\]
\black{
Using the equivalence theorem by \citet{kiefer_wolfowitz_1960}
given in Proposition~\ref{proposition:equivalence}, we have that
%\begin{align}\label{ineq:equivalense_theorem}
$\min_{\lambda \in \triangle(\cX)} \max_{x \in \cX} \| x \|_{M(\lambda)^{-1}}=\sqrt{d}$, which guarantees that 
\[
\max_{x \in \cX'} \|x\|_{M(\lambda^*_{\cX'})^{-1}}=d.
\]
}
\begin{proposition}[\citet{kiefer_wolfowitz_1960}]\label{proposition:equivalence}
Define $M(\lambda)=\E_{z \sim \lambda}[z z^{\top}]$ for any distribution $\lambda$ supported on $\mathcal{X} \subseteq \mathbb{R}^d$. We consider two extremum problems.

The first is to choose $\lambda$ so that 
\[ (1) \lambda \  \text{maximizes} \ \mathrm{det} \ M(\lambda) \hspace{1cm} (\text{D-optimal design})
\]
The second one is to choose $\lambda$ so that 
\[ (2) \lambda \  \text{minimizes} \ \max_{x \in \mathcal{X}}  x^{\top}M(\lambda)^{-1}x \hspace{1cm} (\text{G-optimal design})
\]
We note that $\E_{x\sim \lambda} [x^{\top} M(\lambda)^{-1} x]$ is $d$,
%\wei{It is unclear to me what ``the integral with respect to $\lambda$ of $x^{\top}M(\lambda)^{-1}x$'' means.
%	Is it $\E_{x\sim \lambda} [x^{\top} M(\lambda)^{-1} x]$? If so, perhaps just just this formula.
%Why is it so?}\yuko{I modified it}
hence, $\max_{x \in \mathcal{X}}   x^{\top}M(\lambda)^{-1}x  \geq d$, and thus a sufficient condition for $\lambda$ to satisfy $(2)$ is
\[ (3)  \max_{x \in \mathcal{X}}  x^{\top}M(\lambda)^{-1}x =d.
\]
Statements (1), (2) and (3) are equivalent.
\end{proposition}

This distribution $\lambda^*_{\cX'}$ has the key role for achieving the polynomial-time complexity and optimality of \textsf{PolyABA} in the first phase.

\begin{remark}
$\lambda^*_{\cX'}$ is a $\sqrt{md/\xi_{\min}(\widetilde{M}(\lambda^*_{\cX'}))} \ (\geq 1)$-approximate solution to 
\[\min_{\lambda \in \triangle(\cX)} \max_{x \in \cX} \| x \|_{M(\lambda)^{-1}},\]
where  $m=\max_{x \in \cX} \|x \|_1$ and ${\widetilde{M}({\lambda^*_{\cX'}}})=\sum_{x \in \mathrm{supp}(\lambda^*_{\cX'})}x x^{\top}$.
\end{remark}

Then, based on fixed distribution $\lambda^*_{\cX'}$, we apply static estimation to estimate $\theta$, until we see a big enough gap between the empirically best and the $(d+1)$-th best actions.
Note that computing the empirically best $d+1$ super arms can be done in polynomial time by using \emph{Lawler's k-best procedure}~\citep{Lawler72}.
After the first phase,
we can invoke any BAI-LB algorithms for identifying the best super arm from the empirically best $d$ super arms as a second phase.

If we invoke \textsf{ALBA}~\citep{Tao2018} in the second phase,
 then we have the following theorem.
\begin{theorem}[{{\citet[Theorem~1]{Du+2021}}}]\label{thm:CPE-BL}
	With probability at least $1-\delta$, the \textsf{PolyALBA} algorithm returns the best super arm $x^*$ with sample complexity 
	\begin{align*}
		O\Bigg( & \sum_{i=2}^{\left \lfloor  \frac{d}{2} \right \rfloor} \frac{1}{\Delta_i^2} \left(\ln\frac{|\mathcal{X}|}{\delta}+\ln\ln \Delta_i^{-1}\right) + \frac{d (\alpha \sqrt{m} + \alpha^2 )}{\Delta_{d+1}^2} \left(\ln\frac{|\mathcal{X}|}{\delta}+\ln\ln \Delta_{d+1}^{-1} \right) \Bigg),
	\end{align*}
	where $\alpha = \sqrt{md/\xi_{\min}(\widetilde{M}(\lambda^*_{\cX'}))}$ and $\Delta_i$ denotes the gap of the expected rewards between $x^*$
and the super arm with the $i$-th largest expected reward.
\end{theorem}
%\paragraph{Analysis of the statistical and computational efficiency.}
The first term in Theorem~\ref{thm:CPE-BL} is for the remaining epochs required by subroutine \textsf{ALBA} and
	the second term is for the first phase.
As shown in Theorem~\ref{thm:CPE-BL}, this sample complexity bound has lighter dependence on $\Delta_{\min}$, compared with the existing results by static algorithms~\citep{Kuroki+19}.
Please see \citet{Du+2021} for more detailed discussion on the statistical optimality.
\black{
We note that sample complexity can be improved by choosing a support ${\cX'}$ via E-optimal design since it will minimize the value of $\alpha$. Geometrically, maximizing $\xi_{\min}(\widetilde{M}(\lambda^*_{\cX}))$  can be interpreted as minimizing the diameter of the confidence ellipsoid  (see also Figure~\ref{fig:ellipsoid}). }

\section{Beyond Full-bandit Feedback and Linear Rewards}
%\subsection*{Chapter~\ref{chap:CPE-PL}: Combinatorial Pure Exploration with Partial-Linear Feedback for Nonlinear Rewards}
%\paragraph{Problem and Motivation.}
Although full-bandit settings can capture many practical situations as demonstrated in the previous sections,
it may happen that we cannot always observe outcomes from some of the chosen arms due to privacy concerns or system constraints.
To overcome this issue,
\citet{Du+2021} proposed a general model of \emph{combinatorial pure exploration with partial-linear feedback} (CPE-PL), which simultaneously models limited feedback, general (possibly nonlinear) reward and combinatorial action space.
The model subsumes problems addressed in the previous sections.
In CPE-PL, we are given a combinatorial action space $\cX \subseteq \{0,1\}^d$, where each dimension corresponds to a base arm and each action $x\in \cX$ is an indicator vector of a super arm.
At each round $t$, the agent chooses an action (super arm) $x_t \in \cX$ to pull and observes a random \emph{partial-linear feedback} with expectation of $M_{x_t} \theta$, where $M_{x_t}$ is a transformation matrix $\mathbb{R}^{m_{x_t} \times d}$ whose row dimension $m_{x_t}$ depends on $x_t$ and $\theta \in \mathbb{R}^d$ is an unknown environment vector.
Formally, the feedback vector is written by $y_t=M_{x_t} (\theta + \eta_t ) \in \mathbb{R}^{m_{x_t}}$, where $\eta_t \in \mathbb{R}^d$ is the noise vector. 
%The agent also gains a random (possibly nonlinear) reward related to $x_t$ and $\theta$, which may not be part of the  feedback and thus may not be directly observed.
The CPE-PL framework includes CPE-BL as its important subproblem;
 in CPE-BL, the agent observes full-bandit feedback (i.e., $M_{x_t} = x_t^{\top}$) after each play. 
The model of CPE-PL appears in many practical scenarios, including:
%\begin{spacing}{0.94}
\begin{itemize}
    \item \emph{Learning to rank.} Suppose that a company (agent) wishes to recommend their products to users by presenting the ranked list of items. Collecting data on the relevance of all items might be infeasible, but the relevance of a small subset of items which are highly-ranked (or the top-ranked item) is reasonable to obtain~\citep{Sougara-Ambuj2015,Sougata-Ambuj2016, Sougata-Ambuj2017}.
    In this scenario, the agent selects a ranked list of entire items at each step, and observes random partial-linear feedback on the relevance of highly-ranked $d' \ll d$ items. The objective is to identify the best ranking of their whole items with as few samples as possible.

    \item \emph{Task assignment in crowdsourcing.}
    Suppose that an employer wishes to assign tasks to crowdworkers who can perform them with high quality. It might be costly for the employer and workers to provide task performance feedback for all tasks~\citep{LinTian2014}, and privacy issues may also arise.
    In this scenario, the agent sequentially chooses an assignment of workers to tasks and observes random partial-linear feedback on a small subset of completed tasks.
    The objective is to find the matching between workers and tasks with the highest performance using as few samples as possible.
    
\end{itemize}

We briefly introduce the first polynomial-time algorithmic framework for the general CPE-PL in the fixed confidence setting proposed by~\citet{Du+2021}.
As an important assumption, nonlinear reward functions that satisfy Lipschitz continuity are considered:
\begin{assumption} \label{assumption_Lipschitz}
    There exists a constant $L_p$ such that for any $x \in \cX$ and any $\theta_1, \theta_2 \in \mathbb{R}^d$,
    $|\bar{r}(x, \theta_1)-\bar{r}(x, \theta_2)| \leq L_p ||\theta_1-\theta_2||_2$.
    %\wei{I added the absolute sign around the difference on reward, feeling that this is the standard way.}
\end{assumption}
It is also assumed the exisitence of the \emph{global observer set}:
\begin{assumption} \label{assumption_global}
    There exists a global observer set $\sigma=\{x_1, x_2, \dots, x_{|\sigma|}\} \subseteq \cX$,  such that the stacked $\sum_{i=1}^{|\sigma|} m_{x_i} \times d$ transformation matrix 
    $M_{\sigma}=(M_{x_1}; M_{x_2}; \dots; M_{x_{|\sigma|}})$ is of full column rank ($rank(M_{\sigma})=d$). 
\end{assumption}
% $\sigma=\{x_1, x_2, \dots, x_{|\sigma|}\} \subseteq \cX$,  such that the stacked $\sum_{i=1}^{|\sigma|} m_{x_i}$-by-$d$ transformation matrix $M_{\sigma}=(M_{x_1}; M_{x_2}; \cdots; M_{x_{|\sigma|}})$ is of full column rank, i.e., $\mathrm{rank}(M_{\sigma})=d$. 
The algorithm proposed in~\citet{Du+2021}
samples each super arm in the global observer set to estimate the environment vector $\theta$ and constructs a global confidence bound.
One pull of global observer set $\sigma$ is called \emph{one exploration round}; for the $i$-th exploration round, the agent plays all actions in $\sigma=\{x_1, x_2, \dots, x_{|\sigma|}\}$ once and respectively observes feedback $y_1, y_2, \dots, y_{|\sigma|}$, the stacked vector of which is denoted by $\vec{y}_i=(y_1; y_2; \dots; y_{|\sigma|})$.
Then an estimate of environment vector $\theta$ for the $i$-th exploration round is obtained as  \[\hat{\theta}_i= M_{\sigma}^+ \vec{y}_i,\]
where $M_{\sigma}^+$ is the Moore-Penrose pseudoinverse of $M_{\sigma}$.
From the assumption on the global observer set,
we have $\mathbb{E}[\hat{\theta}_i]=\theta$, i.e., $\hat{\theta}_i$ is an unbiased estimator of $\theta$.
Then, the agent can use the independent estimates in $n$ exploration rounds, i.e.,  \[
\hat{\theta}(n)= \frac{1}{n} \sum_{j=1}^{n} \hat{\theta}_j.\]
Let us define a constant $\beta_{\sigma}$ as follows:
\[
\beta_\sigma:=\max_{\eta_1, \ldots, \eta_{|\sigma|}  \in [-1,1]^{d}}  \left\|  (M_{\sigma}^{\top} M_{\sigma})^{-1} \sum_{i=1}^{|\sigma|} M_{x_i}^{\top} M_{x_i} \eta_i  \right\|_2,\]
which only depends on global observer set $\sigma$.
This $\beta_\sigma$ gives the upper bound on the estimation error of one exploration round;
it holds that
for any $i$,
$\|\hat{\theta}_i -\theta \|_2 \leq \beta_\sigma$~\citep{LinTian2014}.
Based on the upper bound $\beta_\sigma$, a \emph{global confidence radius} is defined as  
\[\textup{rad}_n = \sqrt{ \frac{2 \beta_\sigma^2 \log( 4n^2e^2/ \delta )}{n} }
\] for the estimate $\hat{\theta}(n)$, and it was shown that with high probability, $\textup{rad}_n$ bounds the estimate error of $\hat{\theta}(n)$~\citep{Du+2021}.
 Owing to a global confidence bound and Lipschitz continuity of the expected reward function,
the agent can determine whether the empirically best super arm is indeed the best super arm with confidence $1-\delta$ by simply seeing a large enough gap between the empirically best and second best super arms.
Then, we have the following sample complexity result.
% The experimental results show that our algorithm is superior in terms of speed over the existing ones, and demonstrate that \textsf{GCB-PE} is the first algorithm which can simultaneously deal with combiantorial action space, partial feedback, and nonlinear reward functions.

\begin{theorem}[{{\citet[Theorem~2]{Du+2021}}}]\label{thm:GCB_ub}
	With probability at least $1-\delta$, there exists an algorithm that returns the optimal super arm $x^*$ with sample complexity 
	$$
	O \left( \frac{|\sigma| \beta_\sigma^2 L_p^2}{\Delta_{\textup{min}}^2} \log \left( \frac{ \beta_\sigma^2 L_p^2}{\Delta_{\textup{min}}^2 \delta } \right )   \right ),
	$$
	where $L_p$ is the Lipschitz constant of the reward function.
\end{theorem}

The presented sample complexity heavily depends on the minimum gap $\Delta_{\min}$ due to its static sampling rule.
We remark that it is still open to design an adaptive sampling rule and it is also open to prove a lower bound of the sample complexity for CPE-PL.

\section{Conclusion and Future Directions}

In this article, we reviewed recent advances in combinatorial pure exploration with limited feedback.
Although the combinatorial pure exploration problems can be regarded as an instance of classical linear bandit problems, a naive approach using linear bandit algorithms is computationally infeasible to the problem instance in the combinatorial setting.
We reviewed recently developed polynomial-time algorithms and sample complexity bounds.
These results provided novel insights into online decision making with combinatorial action spaces and combinatorial optimization under uncertainty for incomplete inputs.
Finally, we mention important subjects for the future work.
 
\vspace{0.1cm}

\black{\noindent \textbf{Lower Bounds of Polynomial-time $\delta$-PAC Algorithms.} The optimality of the presented sample complexity bounds for full-bandit settings could be compared with the information theoretic lower bounds for the best arm identification in linear bandits given in Theorem~\ref{thm:lowerbound}:}
\begin{theorem}[{{\citet[Theorem~1]{Fiez2019}}}]\label{thm:lowerbound}
Assume $\eta_t \overset{\mathrm{i.i.d.}}{\sim} \mathcal{N}(0,1)$ for all $t$.
Then for any $\delta \in (0,1)$, $\delta$-PAC algorithm must satisfy
\begin{align*}
	\mathbb{E}_{\theta}[\tau] \geq  \log(1/2.4 \delta) \min_{\lambda \in \triangle(\mathcal{X})}\max_{x\in \mathcal{X} \setminus\{x^*\}} \frac{\|x^*-x\|_{M(\lambda)^{-1}}^2}{((x^*-x)^{\top}\theta)^2}.
\end{align*}
\end{theorem}
As can be seen, there is still a gap between presented sample complexity bounds in Theorems~\ref{thm:topk}, \ref{thm:CPE-BL}, and \ref{thm:GCB_ub} and the information theoretic lower bound in Theorem~\ref{thm:lowerbound}, and we believe that such a gap is needed for reducing the computation cost.
For the nonlinear rewards or partial-linear feedback, no prior work has provided a lower bound even if we consider exponential-time algorithms.
To understand whether or not such a gap is inevitable for the problems of combinatorial pure exploration, it is important to investigate lower bounds of polynomial-time $\delta$-PAC algorithms.
\black{Therefore, we have the following future work.
\begin{problem}
Prove a lower bound of polynomial-time algorithms for combinatorial pure exploration problems with full-bandit or partial-linear feedback.
\end{problem}}

\vspace{0.2cm}
\noindent \black{\textbf{Non-Stationary Setting.} Most existing work focused on the \emph{stationary} case, where the distribution of rewards never changes over time.
However, in real-world applications, we are faced with an extremely \emph{non-stationary} world and it is not reasonable to assume that the distribution stays the same~\citep{besbes2014stochastic}.
For example, in online advertising and recommendation systems,
a user's preferences may likely change when some events happen, which greatly influence the users, and typically exhibit trends on seasonal, weekly, and even daily scales.
Therefore, solving the non-stationary case is promising for increasing the applicability of combinatorial pure exploration methods.
\begin{problem}
Design a method for combinatorial pure exploration problems in non-stationary environments.
\end{problem}}

\section*{Acknowledgements}
YK was supported by Microsoft Research Asia, KAKENHI 18J23034, and JST ACT-X JPMJAX200E.
JH was supported by KAKENHI 18K17998.
MS was supported by KAKENHI 17H00757.

\bibliographystyle{abbrvnat}
\setcitestyle{authoryear,open={((},close={))}}
 \bibliography{mybib.bib} %

% \addcontentsline{toc}{chapter}{Bibliography}
 
% % Non-BibTeX users please use
% \begin{thebibliography}{}
% %
% % and use \bibitem to create references. Consult the Instructions
% % for authors for reference list style.
% % Format for Journal Reference
% Author, Article title, Journal, Volume, page numbers (year)
% Author, Book title, page numbers. Publisher, place (year)
% \end{thebibliography}

\end{document}